# Estimation of Tissue Oxygen Saturation from RGB Images based on Pixel-level Image Translation

Qing-Biao Li[1,3], Xiao-Yun Zhou[1], Jianyu Lin[1,3],Jian-Qing Zheng[1], Neil T. Clancy[2], Daniel S. Elson[1,3]

*[1]The Hamlyn Centre for Robotic Surgery, Imperial College London, London, UK,*

*[2]Centre for Medical Image Computing, University College London, London, UK*

*[3]Department of Surgery and Cancer, Imperial College London, London, UK*
*q.li17@imperial.ac.uk*

## INTRODUCTION

Oxygenation and perfusion reflect tissue metabolic activity, which can be potentially used to monitor diseases including diagnosis or characterisation of cancer. A minimally-invasive endoscopy technique based on HSI uses narrow spectral bands over a continuous spectral range to capture quantitative spectral information for live tissue diagnostics. StO2 can be estimated from HSI images using the Beer-Lambert law, if oxy- and deoxy- haemoglobin are considered to be the primary absorbers within the visible wavelength range, and using an approximation for scattering [1].

The Imperial College London, Structured Light and Hyperspectral Imaging system (ICL SLHSI) is an optical probe system combining sparse hyperspectral measurements and spectrally-encoded structured lighting, which can accurately estimate StO2 with high spatial resolution [2,3]. However, the requirement for an additional optical probe creates a barrier to widespread adoption. In contrast, RGB cameras are widely used during minimally invasive surgery. Hence, it is worthwhile to develop methodologies for estimating the StO2 from RGB images. Currently, HSI images were first recovered from RGB images via 'super-spectral resolution' and then used to estimate StO2 by linear regression (route 1 in Fig. 1)[3], [4].

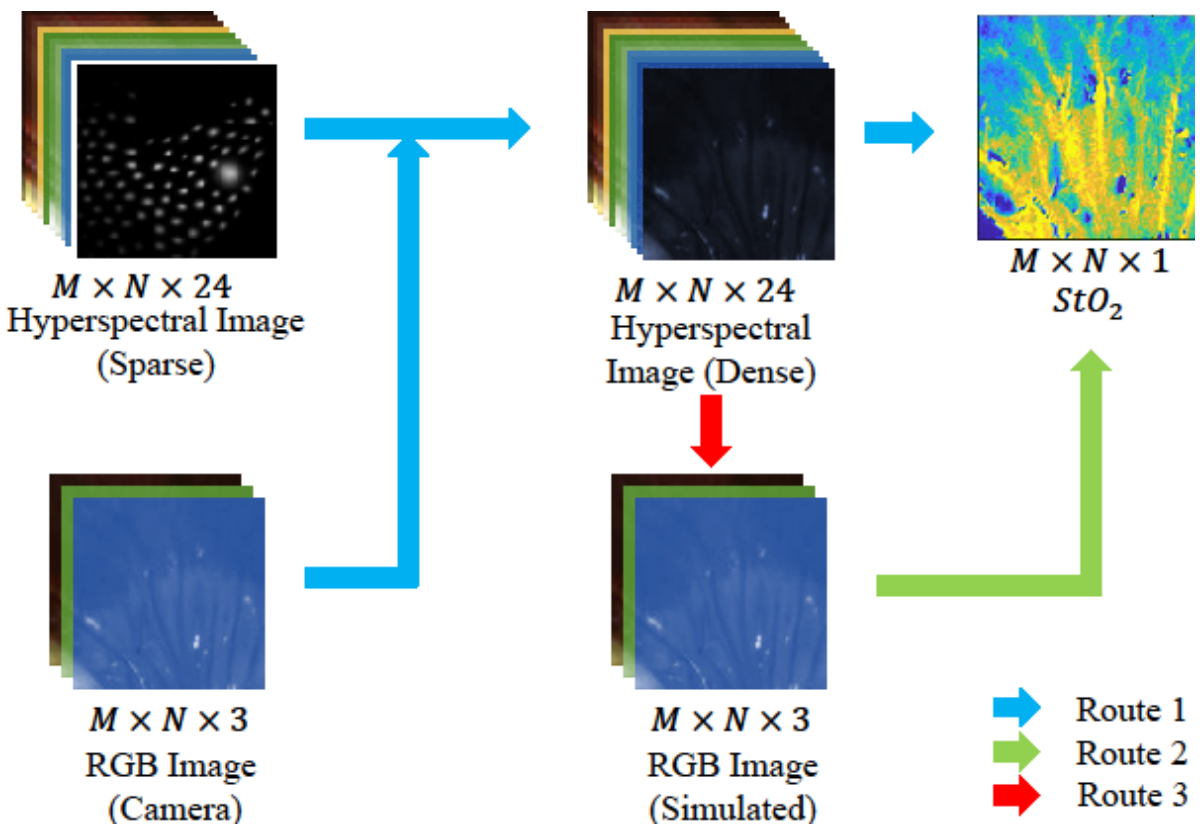

*Fig. 1 Processing methods for StO2 estimation.*

In this paper, conditional Generative Adversarial Networks (cGAN) [5] are introduced to achieve a pixel-level image translation, estimating the StO2 directly from RGB images without super-spectral resolution (route 2 in Fig. 1). The following sections will describe the network architecture, data collection, and validation setup. The test results on three types of *in vivo* data including rabbit uterus, lamb uterus, and porcine bowel are presented and discussed.

## MATERIALS AND METHODS

### Simulation of Dataset

Given an HSI image, an RGB image (input – 'x') can be accurately simulated based on the specific RGB camera responses to the corresponding wavelength range (route 3 in Fig. 1). The intensity ratio $\left(\frac{I_\lambda(x)}{I_{0,\lambda}(x)}\right)$ for each pixel $(i, j)$ reflects the attenuation of the incident light spectrum ( $I_{0,\lambda}(x)$ ) due to oxy- $[HbO_2]$ and deoxy-haemoglobin $[Hb]$ . Given the molar extinction coefficient of each haemoglobin $\varepsilon(\lambda)$ at specific wavelength $\lambda$, the concentration of both haemoglobins can be calculated based on a least squares method [2]:

$$I_\lambda(x) = I_{0,\lambda}(x)e^{-([HbO_2]\varepsilon_{HbO_2}(\lambda)+[Hb]\varepsilon_{Hb}(\lambda)+\alpha)}$$

Where, α is a constant value representing the attenuation caused by scattering and other components within tissue. Thus, the StO2 ground truth can be calculated (image – 'z').

### Conditional Generative Adversarial Networks

In the training stage, the RGB images and the StO2 ground truth image are put into a generator whose structure is shown in Fig. 2, outputting an estimated StO2 (synthesized image – 'y'). Concat(x,y) and Concat(x,z) are separately put into a discriminator, outputting the probability of the input to be z. Here Concat() is concatenate, the probability map is a $30 \times 30 \times 1$ map which is useful for pixel-level rather than image-level translation [5]. The discriminator network architecture is shown in Fig. 3.

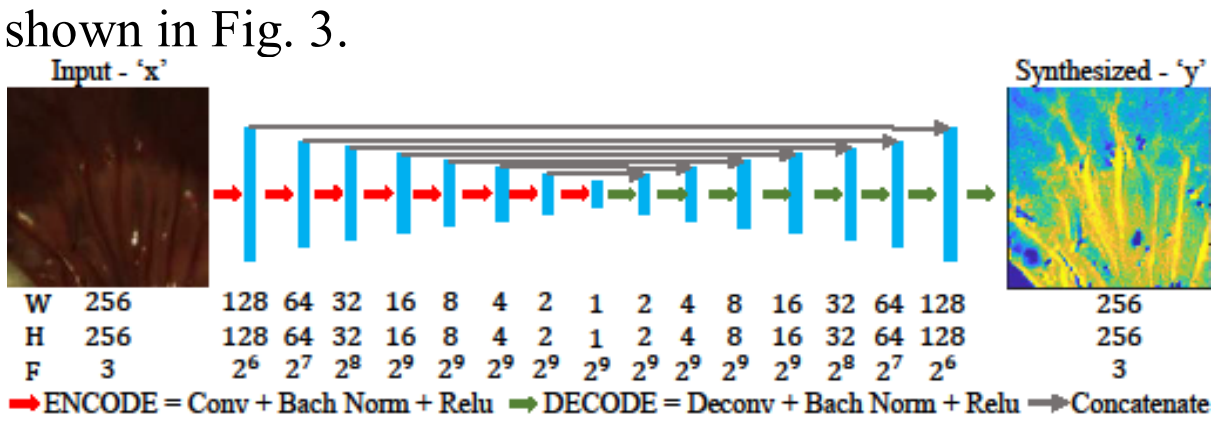

*Fig. 2 Generator network architecture (G).*

Adversarial learning is applied to update the generator and the discriminator models alternately with the loss function.

$$G^* = arg \min_G \max_D L_{cGAN}(G,D) + \lambda L_{L1}(G)$$
$$= arg \min_G \max_D (\log D(y) + \log(1 - D(G(x,z))) + \lambda \|y - G(x,z)\|_1$$

Here, D is the discriminator, G is the generator, $\lambda$ is the L1 weight. The L1 loss $L_{L1}(G)$ is useful to translate the low-frequency information [5].

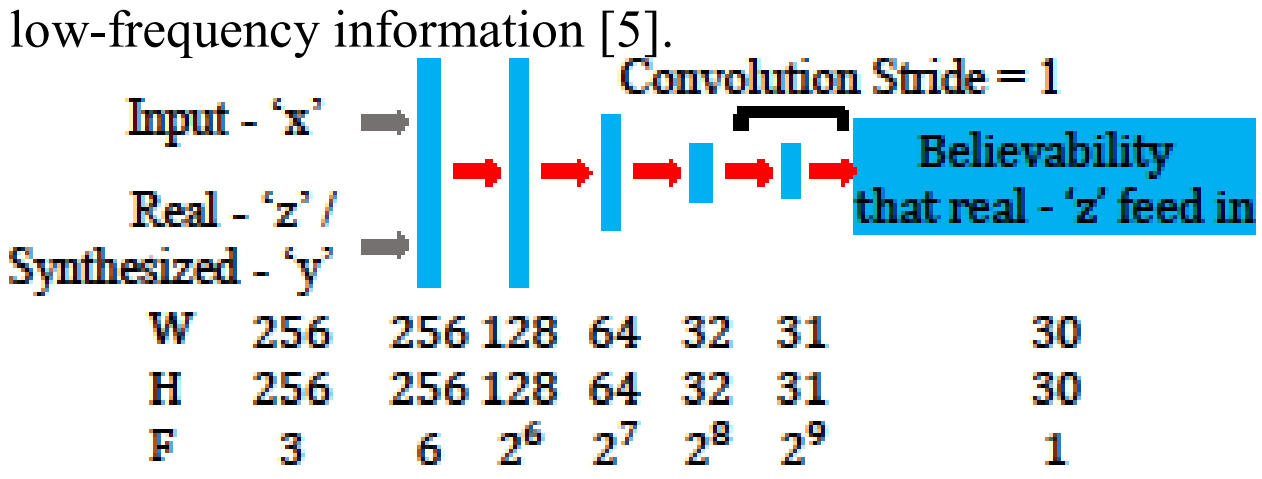

*Fig. 3 Discriminator network architecture (D).*

In the testing stage, the trained generator is used to estimate the StO2 from a new RGB image.

**Validation Setup**

HSI was previously carried out using an LCTF on three kinds of tissues *in vivo* (porcine bowel, lamb uterus and rabbit uterus 222 24-channel hypercubes with spatial sizes 192 × 256 [2]). Then, 167 images were used for training and 55 for testing. The dataset was augmented by slicing images using a 128×128 window with a stride of 16. These cropped images were interpolated to 256×256. The one-channel StO2 was copied to a three-channel image as channel difference between the Input and the Real image may introduce imbalance in the discriminator training. To demonstrate the influence of batch size and L1 weight, the testing was performed for separate cases (one image area cropped from each original image, called intercases) and within one case (the 45 image areas augmented from one original image, called intracases). An Nvidia Titan XP with 12G memory was used for the training. The maximum batch size for the training is 56 while that for the testing is 380. The loss curve usually converged after 10k iterations.

## RESULTS

The mean prediction error between the proposed method and the ground truth was calculated by StO2 intensity difference, which is summarized in Table 1 for different batch sizes with a fixed L1 weight of 100 (left column), and different L1 weights with a fixed batch size of 56 (right column). In intercases we can see that the increase in batch size and L1 weight reduces the inter-case error, which improve the pixel-level translation accuracy. In intracases the trained model was tested in randomly selected tissue images among all its augmented data. The result (Table 1) shows that the increase in L1 weight does not significantly improve the performance. The best model (bold in Table 1) was tested on all images in test set with a batch size of 380, achieving 14.54% in averaged mean error of all tissue images among their augmented data. These errors might be due to indiscernible tissue structures, spectral highlights, bias or noise.

*Table 1. The StO2 prediction error for different setup parameters (Inter. stands for intercases, Intra. Stands for intracases).*

| L1 = 100 | | Batch Size = 56 | |
|---|---|---|---|
| Batch Size | Inter./Intra. Mean Error | L1 Weight | Inter./Intra. Mean Error |
| 1 | 0.1631/- | 50 | 0.0865/0.1103 |
| 16 | 0.1113/- | 100 | 0.0875/0.1151 |
| 32 | 0.0942/- | 200 | 0.0807/0.1292 |
| 56 | 0.0875/- | 400 | **0.0766/0.0841** |

Fig. 4 illustrates a comparison between the StO2 estimated from the synthesized image and the real image, while the intensity difference between them is also presented.

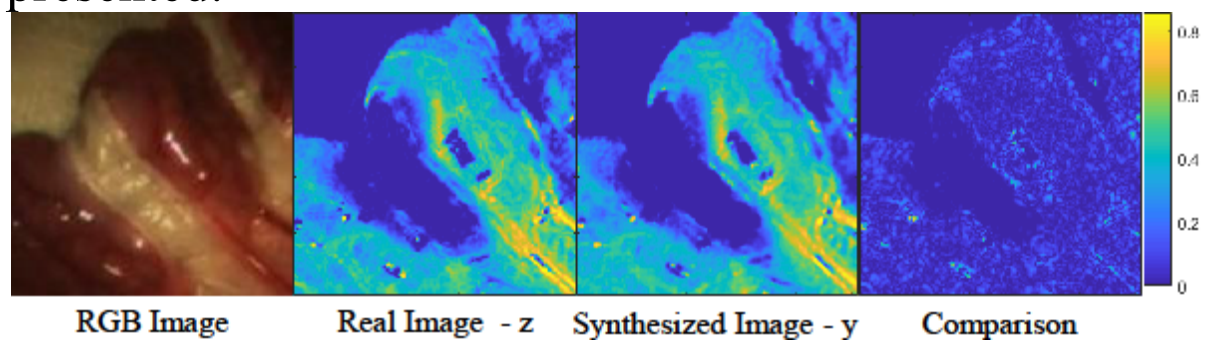

*Fig. 4 Example StO2 images for porcine bowel.*

## DISCUSSION

In this paper, we realized the pixel-level image-to-image translation from RGB images to StO2, with an end-to-end Conditional GAN. The batch size and L1 weight were explored and set to compensate for the inter- and intra-case variations, achieving comparable StO2 predictions. In the future, we will work on the network architecture and the loss function to further improve the accuracy of the estimation as well as analysing the errors.